\RequirePackage{amsmath}
\documentclass[runningheads]{llncs}
\usepackage[colorlinks, citecolor=green]{hyperref}
\usepackage{caption}
\usepackage{orcidlink}
\usepackage[T1]{fontenc}
\usepackage{graphicx}

\usepackage{amsmath}

\usepackage{amssymb}
\usepackage{booktabs}
\usepackage{multirow}
\usepackage{setspace}
\usepackage{booktabs}
\usepackage{hyphenat}

\newcommand{\bftab}{\fontseries{b}\selectfont}

\newcounter{supptable}
\renewcommand{\thesupptable}{S\arabic{supptable}}
\newcommand{\supptable}[1]{%
  \refstepcounter{supptable}%
  \captionsetup{labelformat=empty}%
  \caption{Table \thesupptable: #1}%
}

\newcounter{suppfigure}
\renewcommand{\thesuppfigure}{S\arabic{suppfigure}}
\newcommand{\suppfigure}[1]{%
  \refstepcounter{suppfigure}%
  \captionsetup{labelformat=empty}%
  \caption{Figure \thesuppfigure: #1}%
}

\usepackage{xcolor, colortbl}

\usepackage{cleveref}

\usepackage[acronym]{glossaries}
\loadglsentries{glossary.sty}

\newcommand{\furl}[1]{{\footnotesize\url{#1}}}
\newcommand{\cross}[1][1pt]{\ooalign{%
  \rule[1ex]{1ex}{#1}\cr
  \hss\rule{#1}{.7em}\hss\cr}}

\usepackage{listings}
\lstdefinestyle{mystyle}{
    language=Python,
    basicstyle=\footnotesize\ttfamily,
    keywordstyle=\color{blue},
    commentstyle=\color{green},
    numbers=left,
    numberstyle=\tiny\color{gray},
    stepnumber=1,
    showstringspaces=false,
    tabsize=4,
    breaklines=true,
    breakatwhitespace=false,
    frame=single
}

\begin{document}
%
\title{Mamba? Catch The Hype Or Rethink \\What Really Helps for Image Registration}


\author{
    Bailiang Jian\inst{1,2,3}\textsuperscript{*}\orcidlink{0009-0009-2419-0420},
    Jiazhen Pan\inst{1,2}\textsuperscript{*}\orcidlink{0000-0002-6305-8117},
    Morteza Ghahremani\inst{1,2,3}\orcidlink{0000-0001-6423-6475},
    Daniel Rueckert\inst{1,2,4}\orcidlink{0000-0002-5683-5889},
    Christian Wachinger\inst{1,2,3}\textsuperscript{\cross[.4pt]}\orcidlink{0000-0002-3652-1874}
    \and
    Benedikt Wiestler\inst{1,2}\textsuperscript{\cross[.4pt]}\orcidlink{0000-0002-2963-7772}
}

\def\thefootnote{*}\footnotetext{Equal contribution}
\def\thefootnote{\cross[.4pt]}\footnotetext{Equal Advising}
\def\thefootnote{\arabic{footnote}}

\titlerunning{Catch the Hype or Rethink Registration}

\authorrunning{B. Jian and J. Pan et al.}
%
\institute{
Technical University of Munich, Germany \\
\email{\{bailiang.jian,jiazhen.pan\}@tum.de}
\and
Klinikum Rechts der Isar, Munich, Germany
\and
Munich Center for Machine Learning, Germany
\and
Imperial College London, United Kingdom
}
\maketitle 
\setcounter{footnote}{0}
\begin{abstract}
VoxelMorph, proposed in 2018, utilizes Convolutional Neural Networks (CNNs) to address medical image registration problems. In 2021 TransMorph advanced this approach by replacing CNNs with Attention mechanisms, claiming enhanced performance. More recently, the rise of Mamba with selective state space models has led to MambaMorph, which substituted Attention with Mamba blocks, asserting superior registration. These developments prompt a critical question: does chasing the latest computational trends with ``more advanced'' computational blocks genuinely enhance registration accuracy, or is it merely hype? 
Furthermore, the role of classic high-level registration-specific designs, such as coarse-to-fine pyramid mechanism, correlation calculation, and iterative optimization, warrants scrutiny, particularly in differentiating their influence from the aforementioned low-level computational blocks. In this study, we critically examine these questions through a rigorous evaluation in brain MRI registration. We employed modularized components for each block and ensured unbiased comparisons across all methods and designs to disentangle their effects on performance. Our findings indicate that adopting ``advanced'' computational elements fails to significantly improve registration accuracy. Instead, well-established registration-specific designs offer fair improvements, enhancing results by a marginal 1.5\% over the baseline. 
Our findings emphasize the importance of rigorous, unbiased evaluation and contribution disentanglement of all low- and high-level registration components, rather than simply following the computer vision trends with ``more advanced'' computational blocks. We advocate for simpler yet effective solutions and novel evaluation metrics that go beyond conventional registration accuracy, warranting further research across various organs and modalities.

\end{abstract}
\section{Introduction}
\label{sec:intro}

Efficient and accurate deformable registration is fundamental in neuroimaging analysis, facilitating the understanding of human brain dynamics and the disease course assessment of various neurological disorders~\cite{sotiras2013deformable}. Cross-sectional brain image registration is pivotal when integrating data from a specific cohort to identify unique patterns or when aligning a new patient's scans with population atlases to apply pre-defined anatomical labels~\cite{coffey1992aging,fotenos2005cross-ad,hering2022learn2reg}. 

\paragraph{\textbf{Improvements from low\hyp{}level computational blocks}} 
Accurate, efficient, and rapid image registration remains a primary goal, both for research and clinical application. Over the past decade, numerous deep learning\hyp{}based image registration methods have been proposed. Convolutional Neural Networks (CNNs), such as Voxelmorph\hyp{}based methods~\cite{balakrishnan2019voxelmorph,jian2022weakly}, pioneered the field of learning\hyp{}based registration methods by utilizing learned local convolutional priors to enhance registration accuracy and accelerate the inference process. Lately, Vision Transformer\hyp{}based methods~\cite{chen2022transmorph,ghahremani2024hvit,meng2023nice-trans}, such as Transmorph, have been introduced. These methods leverage the inherent attention mechanism to ensure global dependencies and registration coherence. Although these models claim higher registration accuracy than Voxelmorph, their inference speed is hindered by the quadratic computational complexity of the attention mechanism. Recently, Mamba~\cite{gu2024mamba}, utilizing Selective State Space Models (SSMs), has been introduced in registration~\cite{guo2024mambamorph}, claiming to balance both accuracy and efficiency by capturing long\hyp{}range dependencies with nearly linear computational complexity. However, in our view, more experimental evidence with fair comparisons (same settings, training strategy, etc.) and more datasets are required to substantiate this argument.

\paragraph{\textbf{Improvement from high\hyp{}level registration\hyp{}specific designs}} In contrast to low\hyp{}level computational blocks which enhance the registration performance at the lowest architecture level in a ``brute force'' alike way without clear justification or motivation, registration\hyp{}specific designs address the problem at a higher architecture level considering the characteristics of 
 registration tasks. \textbf{Coarse\hyp{}to\hyp{}fine optimization pyramid} \cite{rueckert1999nonrigid,brox2004high,ghahremani2024hvit,sun2018pwc} is a widely\hyp{}used and effective technique that first estimates large deformations at a coarse scale and refines smaller motions at a finer scale. This approach is often combined with \textbf{iterative warping optimizations} \cite{rueckert1999nonrigid,pan2021efficient,teed2020raft,zhao2019recursive}, where large deformations are approximated through multiple estimation steps, allowing for fine\hyp{}tuning during the registration process. Additionally, unlike U\hyp{}Net\hyp{}like architectures that concatenate fixed and moving images before feeding them into networks, the \textbf{dual\hyp{}branch} approach with two separate encoders has gained popularity in registration and optical flow estimation~\cite{kang2022prnet,meng2022nice,sun2018pwc,teed2020raft}. This method can explicitly distinguish the dissimilarity between two images. Finally, calculating the regional \textbf{correlations} between fixed and moving images to establish voxel\hyp{}wise correspondences~\cite{hosni2012fast,kang2022prnet,meng2024corr-wmlp,pan2022learning} is vital and beneficial for accurate image registration.\\
\\
\noindent As academic researchers specializing in medical image registration for several years, we have observed a recurring trend: the adaptation of low-level computation blocks, such as Transformers, Multilayer Perceptrons (MLPs), and Mamba, often without clear justification and motivation in the core of the registration problem. These methods have garnered significant attention. On the contrary, registration-specific designs have gained much less attention. We argue that techniques like coarse-to-fine approaches, correlation and iterative optimization hold substantial potential for achieving better results. Moreover, registration studies are typically proposed with multiple novel components, wherein advanced low-level computational blocks and registration-specific designs are \textbf{intertwined}. The performance contributions of individual components within these complex models, under controlled settings (same data domain and training presets), remain unclear.

\noindent The major contributions of our work can be summarized as follows:
\begin{enumerate}
    \item \textbf{Modularized Component Analysis}: We conducted a comprehensive study using modularized components for every aforementioned block and performed fair comparisons across all blocks. This approach allowed us to disentangle, isolate, and identify the contribution of each component to the final registration results, providing clarity on which elements are most impactful.

    \item\textbf{Extensive Dataset Evaluation}: Our experiments were carried out on five open-source brain MRI datasets with consistent experimental settings for all runs. This ensured a fair and controlled evaluation of Transformers, Mamba, and CNN-based methods, enabling a reliable performance comparison. We demonstrate the qualitative results with random selections to avoid any bias. 

    \item\textbf{Recommendation}: Our findings reveal that ``advanced'' low-level computational blocks (Transformers and Mamba) offer worse results compared to CNN-based methods in Brain MRI registration tasks. The primary contributions to registration performance stem from high-level registration-specific designs such as correlation layers and iterative warping pyramids. Based on these insights, we encourage the community to focus more on registration-specific designs instead of simply transplanting the ``advanced'' computational blocks without considering the registration characteristics. We advocate for focusing on simpler, more effective solutions and novel evaluation metrics that move beyond the conventional registration accuracy score, as our results reveal nuanced differences among the tested models.

\end{enumerate}


\section{Methodology}\label{sec:method}

\textbf{Preliminaries} 
Given a target image and a source image with their anatomical labels, $(I_t,L_t)$ and $(I_s,L_s)$, respectively, we want to build a network parametrized by $\theta$ to predict a dense deformation field $\phi$ which establishes the voxel-wise spatial correspondence between target and source images $F_\theta(I_s, I_t)=\phi$. 
To optimize the parameters $\theta$, we define the training loss as:
\begin{equation}
    \mathcal{L} = \mathcal{L}_\textrm{sim}(I_t, I_s\circ\phi) + \gamma\mathcal{L}_\textrm{seg}(L_t, L_s\circ\phi) + \lambda\mathcal{L}_\textrm{reg}(\phi),
\end{equation}
where $I_s\circ\phi$ represents the warped source image, $\mathcal{L}_\textrm{sim}$ measures the similarity between images, $\mathcal{L}_\textrm{seg}$ computes the dice loss between segmentation labels, and $\mathcal{L}_\textrm{reg}$ is the smoothness regularization function which penalizes the irregularities of $\phi$. $\gamma$ and $\lambda$ are pre-determined hyperparameters for Dice loss and regularization, respectively. 

We apply VoxelMorph~\cite{balakrishnan2019voxelmorph} as our registration vanilla method to model $\theta$ and estimate the deformation field $\phi$. It utilizes U-Net~\cite{ronneberger2015u} as the architecture backbone and takes concatenated target and source images as the network inputs. U-Net forward-passes them to the encoders with CNN-based low-level computational blocks and reduces the image size when reaching the U-Net bottleneck, and upsamples to the original image size in its decoders. It should be noted that instead of using max-pooling and trilinear upsampling to down-/upsample the feature maps, we apply convolutional kernels with stride size 2 and transposed convolutional kernels for down-/upsampling to boost the networks' performance.

\subsection{Improvement using advanced low-level computational blocks}

\paragraph{\textbf{Vision Transformer}}~\cite{dosovitskiy2020image}, introduced in computer vision in 2021, has demonstrated superior performance over CNN-based methods~\cite{he2016deep} in various tasks, such as image classification, detection, and segmentation. It leverages Attention mechanisms that can model global dependencies, unlike local kernels used in CNNs. Based on that, Swin-Transformer was proposed~\cite{liu2021swin}, using the shift-window mechanism and achieved linear computational complexity with respect to image size. Swin-Transformer was further introduced into the image registration field~\cite{chen2022transmorph}. In TransMorph, the authors reported improved registration performance and more coherent deformation fields compared to VoxelMorph due to leveraging these global dependencies. In our work,  we replace the computational building blocks (CNNs in the baseline encoder) with Transformers to attempt to improve the registration performance, and this network is dubbed as ``TM''.

\paragraph{\textbf{Mamba}}~\cite{gu2024mamba} improves the efficiency of State Space Models (SSMs), which are linear time-invariant systems that map stimulation to response with linear ordinary differential equations. 
The continuous-time model is first discretized to enable integration into deep learning algorithms. Further, Mamba introduces a selective scan mechanism and hardware-accelerated algorithm, incorporating time-dependent information and significantly boosting SSM efficiency. 
Concurrently introduced to the registration field, Mamba has shown strong capabilities in handling long sequences with nearly linear complexity~\cite{guo2024mambamorph}. In this work, we attempt to improve the registration accuracy by replacing the computational elements from CNNs with Mamba blocks. This replaced network version is denoted as ``Mam''.

\paragraph{\textbf{Large-Kernel Convolution}}~\cite{ding2022replk} shows the capability of long-range dependencies as well by scaling up the kernel size. LKU-Net~\cite{jia2022lku} claimed that the effective receptive field of a parallel convolutional block with the largest kernel size of 5 is sufficient to address brain MRI registration without global dependencies. They achieve similar dice score as TransMorph while with more implausible deformations. In this work, we evaluate the influence of replacing computational elements from CNNs with large-kernel CNNs and denote it as ``LKU''.

\begin{figure*}[ht]\label{f_main}
    \centering
    \includegraphics[width=1.0\linewidth]{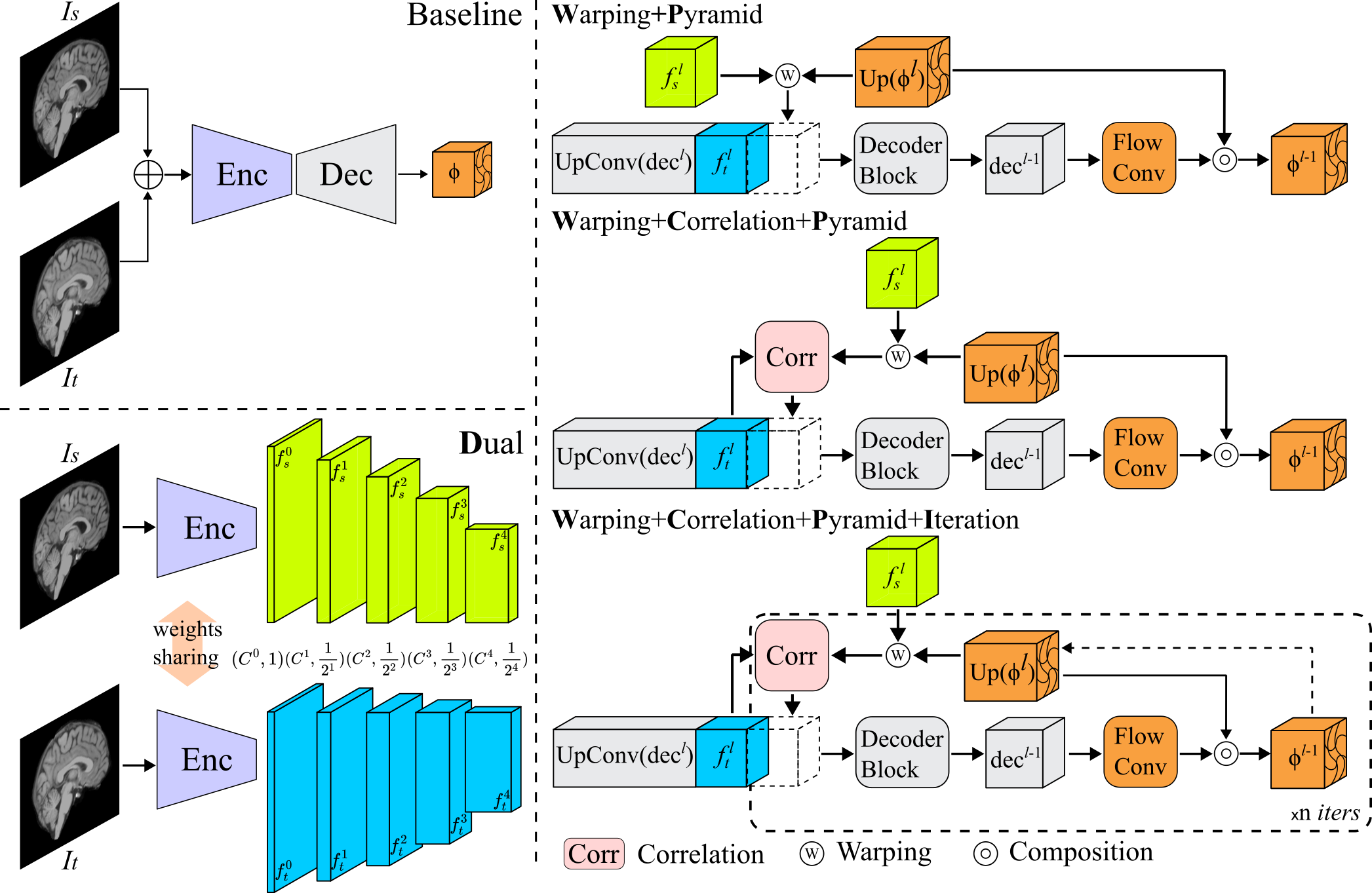}
    \caption{Overview of the baseline and the modularized components. (a) The upper left shows the U-Net-based \textit{Baseline} methods, which concatenates the source and target images and predicts the final deformation field directly. (b) The bottom left shows the \textit{\textbf{D}ual}-stream encoder variant. (c) The right figure presents the detailed workflow of each registration-specific element (\textit{\textbf{P}yramid, \textbf{W}arping}, \textit{\textbf{C}orrelation} and \textit{\textbf{I}teration}) at the specific level $\ell$. Level $\ell$ corresponds to $2^{-\ell}$ resolution. The width of the cubes in the right diagram does not correspond to the exact number of channels of the features, but only for display convenience.
    }
    \label{fig:arch}
\vspace{-0.1cm}
\end{figure*}

\subsection{Improvement using high-level registration-specific designs}
\paragraph{\textbf{Dual Stream Encoders}} Dual stream encoders utilize two separate branches with shared weights to encode the target and source images independently~\cite{meng2022nice,shi2022xmorpher}. Compared to the widely-applied U-Net-based architectures~\cite{balakrishnan2019voxelmorph,chen2022transmorph,guo2024mambamorph}, which concatenate the two images before feeding them into the encoder, dual stream encoders are favored in modern optical flow works~\cite{sun2018pwc,teed2020raft}. This approach constrains the network to produce meaningful representations of each image separately before combining them at a higher level. This process resembles the standard matching approach, akin to standard matching process where features are extracted from image patches and then compared. Additionally, the dual stream encoder architecture facilitates the straightforward introduction of motion pyramid, warping, and correlation layers (see the following paragraphs for more details). This technique is denoted as ``\textbf{D}'' in this work.

\paragraph{\textbf{Motion Pyramid and Warping}} The Motion Pyramid architecture, also known as coarse-to-fine deformation field estimation, is a classic and widely used technique for motion tracking and registration~\cite{rueckert1999nonrigid,brox2004high,ghoul2024attention,sun2018pwc}. In registration networks, a pyramid of feature maps of the target and source images are initially generated. The feature maps at a coarse scale (typically $1/8$ or $1/16$) are used to predict global and large deformations. The finer moving feature is then warped using this estimated deformation to reduce dissimilarity. By estimating the motions bottom up, the deformation field are progressively refined at finer scales. We denote the motion pyramid and warping as ``\textbf{PW}'' in this work.

\paragraph{\textbf{Correlation Layers}} Correlation layers can be used in registration to explicitly measure region dissimilarity and guide deformation field estimation~\cite{dosovitskiy2015flownet,kang2022prnet,pan2024reconstruction,teed2020raft}. In this work, we first flatten the spatial dimensions of the feature maps to obtain feature vectors ${\mathbf{f}_t,\mathbf{f}_s}\in\mathbb{R}^{HWD\times N}$, where $N$ denotes the embedding feature length. We then compute the correlation as the inner product between these two feature vectors:
\begin{equation}
    \mathbf{C} = \textrm{Corr}(\mathbf{f}_t,\mathbf{f}_s)=\frac{1}{N}\mathbf{f}_t^{\top} \mathbf{f}_s,
\end{equation}
resulting in $\mathbf{C}\in\mathbb{R}^{HWD\times HWD}$. To reduce the computational complexity, we compute a partial correlation within a neighborhood volume of size $d^3$ and indicate the correlation layers as ``\textbf{C}'' in this study. 

\paragraph{\textbf{Iterative optimization}} is applied in many recent works to improve the registration accuracy~\cite{ma2024iirp,pan2021efficient,qiu2022embedding,teed2020raft,zhao2019recursive}. The direct deformation prediction is decomposed by the iterative refinement in which the registered image is optimized step by step. We apply this registration optimization strategy in this work, and the exact architecture is depicted in~\Cref{fig:arch}. This technique is abbreviated as ``\textbf{I}''.

\section{Experiments and Results}
\label{sec:experiment}
\begin{table}
\caption{The architectural details of each method. We report the number of trainable parameters (Params) in the unit of million (M), the number of floating point operations (FLOPs) in the unit of GB, the maximum allocated GPU memory during training (Max. Mem Train) in the unit of GB, and the GPU runtime of inference on a pair of images (Runtime Infer) in the unit of second.}\label{tab:net_param}
    \centering
    \setlength{\tabcolsep}{2.5pt} 
\begin{tabular}{lcccc}
    \toprule
       &  Params & FLOPs & Max.Mem & Runtime\\
       &    (M)   & (GB) &  Train (GB) & Infer (s)\\\midrule
    VXM & 2.6 & 235.2 & 8.54 & 5.27\\
    Mam-VXM & 2.4 & 231.4 & 10.86 & 7.66\\
    TM & 46.6 & 379.1 & 9.22 & 5.09 \\
    Mam-TM & 39.0 & 289.1 & 9.63 & 5.44 \\
    LKU & 8.3 & 517.4 & 10.74 & 3.68 \\
    \midrule
    Dual & 1.9 & 216.5 & 8.55 & 6.76 \\
    DWP & 1.9 & 169.5 & 8.59 & 6.31 \\
    DWCP & 8.4 & 214.6 & 9.12 & 9.94 \\
    DWCPI & 8.1 & 380.1 & 9.56 & 12.42 \\
    \bottomrule
\end{tabular}
\end{table}


\paragraph{\textbf{Datasets}} We use OASIS~\cite{marcus2007oasis}, ADNI~\cite{jack2008adni}, IXI\footnote{\url{https://brain-development.org/ixi-dataset/}}, LPBA~\cite{shattuck2008lpba}, and Mindboggle~\cite{klein2005mindboggle} to conduct experiments on mono-modal cross-sectional brain MRI deformable registration. The train/test split are as follows: OASIS (330/84), ADNI (234/43), IXI (301/115), LPBA (0/40, zero-shot), Mindboggle (0/100, zero-shot). During training, random pairs are taken within each datasets. Importantly, using identical random seed and dataloader settings, we ensure that all methods register the same pair of images in each training iteration. For testing, 200 separate random pairs are sampled from each dataset.

\paragraph{\textbf{Evaluation metrics}} We compute the Dice score (DSC) and the 90th percentile of the Hausdorff distance (HD90) using the anatomical segmentation labels as metrics of registration accuracy. To assess the deformation field plausibility and smoothness, we report the standard deviation of the logarithm of the Jacobian determinant (SDlogJ) and the percentage of non-diffeomorphic voxels (NDV)~\cite{liu2024ndv} within the brain area.

\paragraph{\textbf{General Setup}} 
The layer to predict the flow field(s) at any level in any methods, i.e., the \textit{FlowConv}, is a $k=3, s=1, out\_channel=3$ convolution with weights initialized from normal distribution (0, $10^{-5}$) and bias initialized as zero. The final predicted displacement field of each method is at half-resolution. It is then upsampled and upscaled to warp the full-size source image and label. The weights $\gamma,\lambda$ of $\mathcal{L}_\textrm{seg}$ and $\mathcal{L}_\textrm{reg}$ are set to 0.5. For the motion pyramid, at each sub-level, we only compute the image similarity and smoothness regularization. The losses are scaled in proportion to the resolution of the output level. For example, at level $4$ where the resolution is $2^{-4}$, the computed loss is $2^{-4} * (\mathcal{L}_\textrm{sim} + \lambda\mathcal{L}_\textrm{reg})$. The motion pyramid is computed at resolutions [1/16, 1/8, 1/4]. The random seed of training and testing is set to 2023. All methods are trained with 100 epochs with learning rate $10^{-4}$ with exponential decay rate $0.996$.

\paragraph{\textbf{Method-Wise Setup}}
For baseline methods, the channel number of the vanilla \textit{VoxelMorph (VXM)} model is set to [16,32,64,96,128], [128,96,64,32] respectively. There are two remaining convs with 32 channels before \textit{FlowConv}. \textit{Mam-VXM} replaces the CNNs in VXM's encoder with Mamba blocks while keeping the rest identical. \textit{TransMorph (TM)} uses $embed\_dim=96$ as referred in the paper~\cite{chen2022transmorph}. Similar to \textit{Mam-VXM}, \textit{Mam-TM} replace the Swin-Transformers in TM's encoder with Mamba blocks. \textit{LKU-Net} use $large\_kernel\_size=5, feat\_chan=16$ as referred in the paper~\cite{jia2022lku}. The \textit{\textbf{D}ual}-stream variant of VXM halves the encoder feature channel numbers, while the motion \textit{\textbf{P}yramid} and \textit{\textbf{W}arping} variant's decoder is using the same number of channels as VXM without the remaining convolutions. The \textit{\textbf{C}orrelation} variant computes correlation with different number of radius at different level. At the bottom $1/16$ resolution, the global correlation is computed, while radii of [3,2,1] are used at the rest resolutions [1/8,1/4,1/2]. From the preliminary experiment, we found that only iterations at the last two levels will contribute to performance improvement. Therefore, the \textit{\textbf{I}teration} variant sets number of iterations $n=2$ at resolution [1/4, 1/2]. The numerical architecture details can be found in~\Cref{tab:net_param}.  

\begin{table}[ht]
\caption{Cross-sectional registration results of brain T1-MRI. Separate 200 pairs are randomly sampled from 100 subjects in Mindboggle and 40 subjects in LPBA. SD$\log$J is in the scale $\times10^2$. The top two results are marked in bold.}\label{tab:lpba_mindboggle}
\scriptsize
\centering
\setlength{\tabcolsep}{0.26mm}{}
\begin{tabular}{lcccccccc}
    \toprule
    & \multicolumn{4}{c}{LPBA}
    & \multicolumn{4}{c}{MindBoggle}
    \\
    \cmidrule(lr){2-5} \cmidrule(lr){6-9}
       &  DSC $\uparrow$ & HD90 $\downarrow$ & SD$\log$J $\downarrow$ & NDV(\%) $\downarrow$ &  DSC $\uparrow$ & HD90 $\downarrow$ & SD$\log$J $\downarrow$ & NDV(\%) $\downarrow$\\
    \cmidrule(lr){2-5} \cmidrule(lr){6-9}
    affine &54.0$\pm$4.8& - & - & - &51.5$\pm$4.3& - & - & - \\
VXM & 67.0$\pm$3.6 & 7.54$\pm$0.81 & 7.71$\pm$0.79 &0.35$\pm$0.14 &68.8$\pm$2.3 & 5.41$\pm$0.73 & 8.51$\pm$0.48 &0.34$\pm$0.07 \\
Mam-VXM & 67.5$\pm$3.3 & 7.45$\pm$0.76 & 7.93$\pm$0.88 &0.40$\pm$0.10 &69.2$\pm$2.3 & 5.37$\pm$0.69 & 9.07$\pm$0.54 &0.45$\pm$0.09 \\
TM & 67.3$\pm$3.3 & 7.43$\pm$0.75 & 7.77$\pm$0.81 &0.35$\pm$0.10 &68.8$\pm$2.3 & 5.35$\pm$0.70 & 8.94$\pm$0.51 &0.40$\pm$0.07 \\
Mam-TM & 66.8$\pm$3.5 & 7.53$\pm$0.77 & 7.92$\pm$0.80 &0.37$\pm$0.10 &68.6$\pm$2.3 & 5.37$\pm$0.70 & 8.88$\pm$0.54 &0.39$\pm$0.08 \\
LKU & 67.2$\pm$3.2 & 7.42$\pm$0.73 & 8.43$\pm$1.31 &0.61$\pm$0.38 &68.8$\pm$2.2 & 5.35$\pm$0.69 & 11.61$\pm$1.50 &0.85$\pm$0.16 \\
Dual & 66.4$\pm$3.7 & 7.57$\pm$0.79 & 8.43$\pm$1.11 &0.44$\pm$0.25 &68.6$\pm$2.2 & 5.38$\pm$0.71 & 8.69$\pm$0.51 &0.42$\pm$0.08 \\
DWP & 70.4$\pm$2.6 & 6.91$\pm$0.60 & 7.19$\pm$0.64 &0.24$\pm$0.06 &69.4$\pm$2.2 & 5.33$\pm$0.69 & 8.47$\pm$0.52 &0.35$\pm$0.06 \\
DWCP & \bftab{70.4$\pm$2.7} & \bftab{6.98$\pm$0.64} & \bftab{6.59$\pm$0.71} & \bftab{0.07$\pm$0.05} & \bftab{69.6$\pm$2.2} & \bftab{5.28$\pm$0.69} & \bftab{7.47$\pm$0.40} & \bftab{0.08$\pm$0.02} \\
DWCPI & \bftab{71.3$\pm$2.3} & \bftab{6.88$\pm$0.59} & \bftab{6.82$\pm$0.43} & \bftab{0.01$\pm$0.01} & \bftab{69.8$\pm$2.2} & \bftab{5.28$\pm$0.69} & \bftab{7.68$\pm$0.42} & \bftab{0.01$\pm$0.00} \\
    
    \bottomrule
\end{tabular}
\end{table}

\begin{figure}[ht!]
    \centering
    \includegraphics[width=\linewidth]{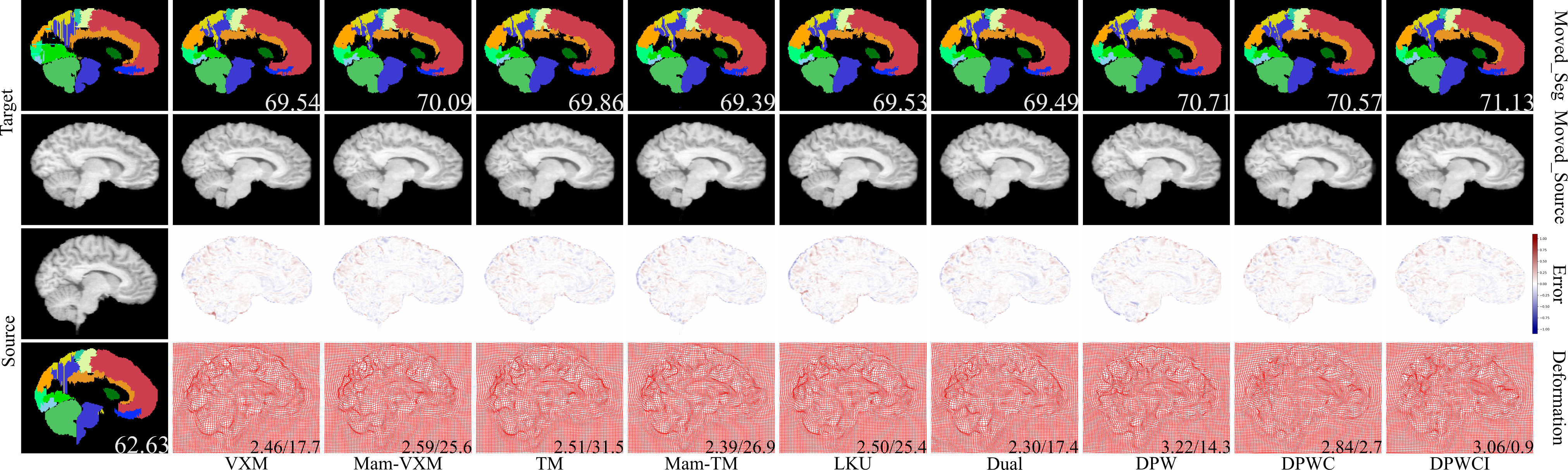}
    \caption{
    The sagittal-viewed visualization results of a randomly sampled pair from the LPBA dataset. The first column shows the target label map and image, and the source image and label map from top to bottom. The other columns correspond to the methods to be compared. The first and second rows display the warped source (moved) label map and image by respective methods. The bottom right number shown in the moved label map is the mean dice score (DSC) of the volume (not the slice). The third row depicts the subtraction map (error map) between the target and moved images. The value is within the range [-1,1] since the image intensities are normalized to [0,1]. The last row shows the warped grid of the deformation field. The two numbers shown are the mean foreground displacement magnitude and the percentage of foreground non-diffeomorphic voxels, both statistics are computed for the entire image volume.
    }
    \label{fig:lpba}
\vspace{-0.1cm}
\end{figure}

\begin{figure}[ht!]
    \centering
    \includegraphics[width=\linewidth]{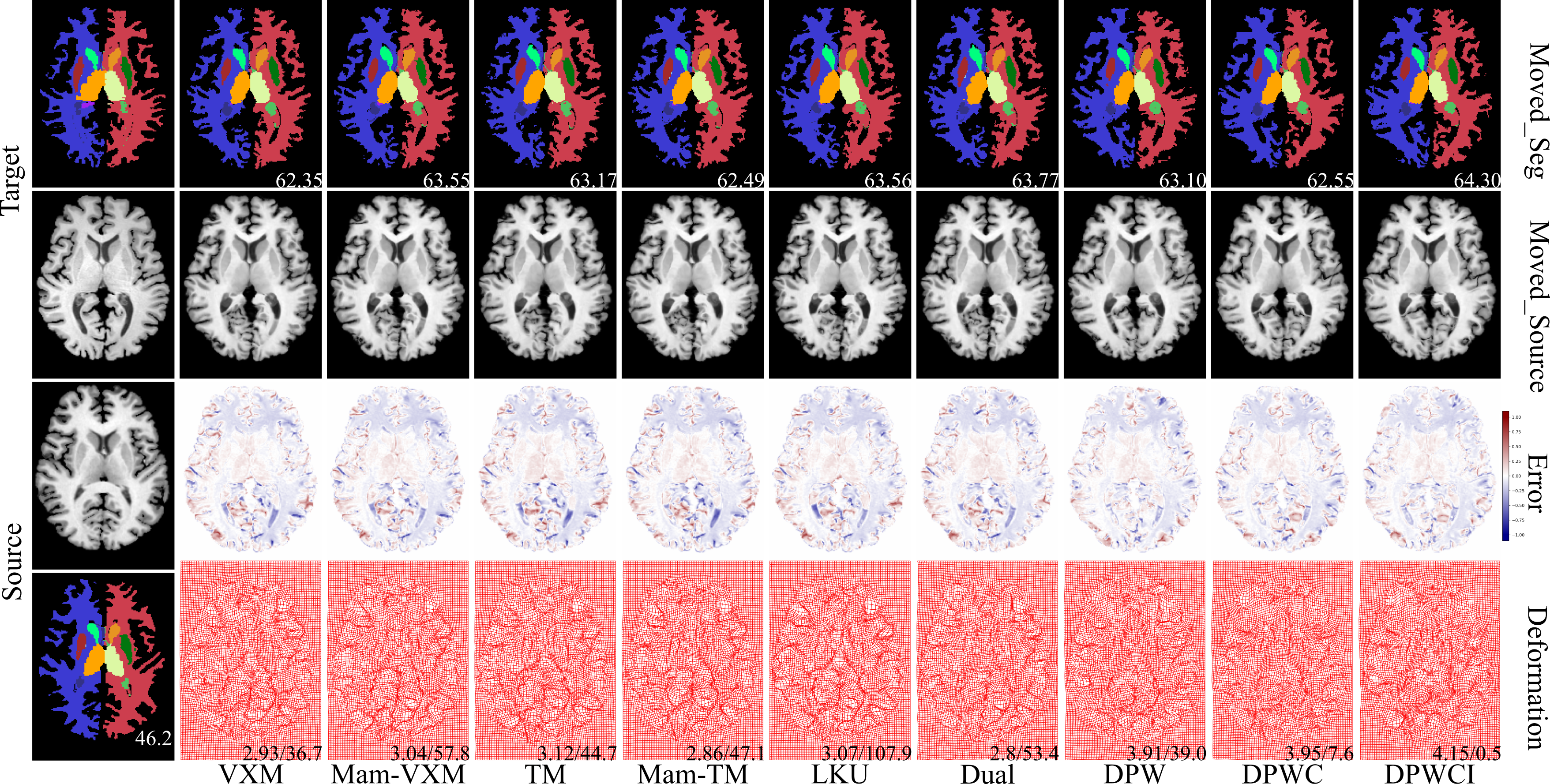}
    \caption{
    The axial-viewed visualization result of a randomly sampled pair from the Mindboggle dataset. The first column shows the target label map and image, and the source image and label map from top to bottom. The other columns correspond to the methods to be compared. The first and second rows show the warped source (moved) label map and image by respective methods. The bottom right number shown in the moved label map is the mean dice score (DSC) of the volume (not the slice). The third row depicts the subtraction map (error map) between the target and moved images. The value is within the range [-1,1] since the image intensities are normalized to [0,1]. The last row shows the warped grid of the deformation field. The two numbers shown are the mean foreground displacement magnitude and the percentage of foreground non-diffeomorphic voxels, both statistics are computed for the entire image volume.
    }
    \label{fig:mindboggle}
\vspace{-0.1cm}
\end{figure}

\paragraph{\textbf{Results}} The zero-shot evaluation on LPBA and MindBoggle, with quantitative results, is presented in~\Cref{tab:lpba_mindboggle}. The qualitative visualizations on LPBA and MindBoggle are shown in~\Cref{fig:lpba} and~\ref{fig:mindboggle}. Comprehensive qualitative (sagittal and axial views) and quantitative evaluations on OASIS, ANDI, and IXI are available in the \textbf{Supplementary materials}. In a controlled experimental environment, it is observed across all datasets that using more advanced computational blocks, such as Mamba or Transformers, does not improve and may even worsen performance. In contrast, utilizing registration-specific designs on top of VXM\footnote{All the registration-specific designs is only applied on top of VXM in this work.} can enhance registration performance, with improvements of approximately 1.5\% and 5\% in LPBA's zero-shot evaluation. Additionally, a qualitative evaluation was performed on randomly selected subjects from all five datasets. The differences in registered images, segmentations, residual errors, and deformation fields between different methods are subtle across all datasets, with the registration performance of VXM (baseline method) already being satisfactory.

\section{Discussion and Conclusion}
\label{sec:conclu_discuss}

In this work, we set out to answer two critical questions: 1. Do ``more advanced'' low-level computational blocks, such as Transformer or Mamba, genuinely enhance medical imaging registration? 2. If ``chasing trends'' does not lead to substantial improvements, what truly can enhance medical image registration? Through comprehensive and rigorous experiments on five brain MR datasets, our findings revealed that employing ``more advanced'' computational blocks can barely improve, and in most cases, even worsen, registration accuracy. In contrast, incorporating high-level registration-specific designs, such as coarse-to-fine and correlation layers, can improve registration performance, albeit marginally, with a 1.5\% increase in dice score compared to the baseline. Notably, the vanilla \textit{Voxelmorph} already produced competitive qualitative registration results, nearly indistinguishable from the best-performing models. Therefore, we encourage the community to apply the simplest \textit{VoxelMorph} to handle the most brain image registration problems, and to use \textit{DPWCI} networks in the cases if the marginal higher registration accuracy is demanded. We release all these models and relevant codes at \url{github.com/rethink-reg}.

Based on our findings, we advise the community to approach the development of complex, trend-driven registration architectures \textbf{with caution}. Simplifying designs can often yield comparable or even superior results. While the pursuit of innovative techniques is essential, it is equally crucial to understand why a particular method is suitable for registration and what factors contribute to its success. We also advocate for the development of new evaluation metrics, moving beyond the conventional registration accuracy score, as our results show nuanced differences among the tested models. Moreover, we suggest focusing on more substantive and impactful areas, including real-time and data-efficient registration, robustness, generalizability, patient-specific adaptation, and the interpretability of deep learning models.

\begin{credits}
\subsubsection{\ackname}
This work was supported by BMWi (project ``NeuroTEMP'') research funding and the Munich Center of Machine Learning (MCML). We gratefully thank the review and feedback provided by Yawei Li.
\subsubsection{\discintname}
The authors declare no competing interests.
\end{credits}

\bibliographystyle{splncs04}
\bibliography{main}
\clearpage
\appendix
\section{Supplementary Materials}

\begin{figure}[ht]
    \centering
    \includegraphics[width=\linewidth]{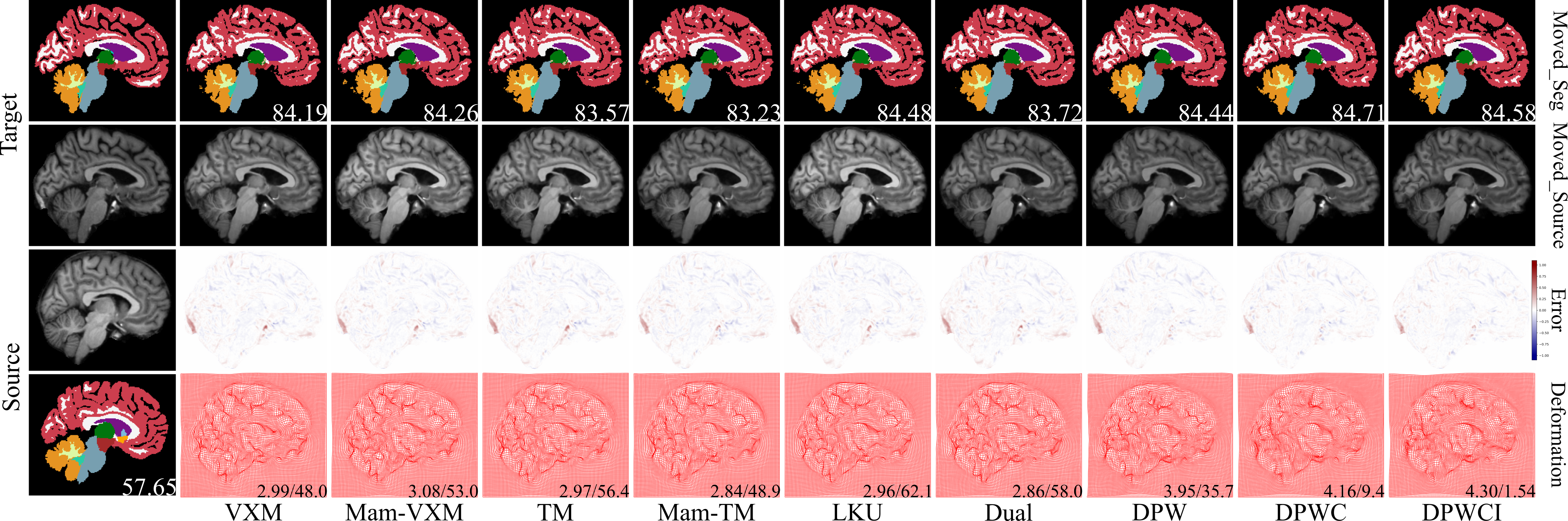}
    \suppfigure{
    The sagittal-viewed visualization result of a randomly sampled pair from the OASIS dataset. The first column shows the target label map and image, and the source image and label map from top to bottom. The other columns correspond to the methods to be compared. The first and second rows show the warped source (moved) label map and image by the methods. The bottom right number shown in the moved label map is the mean dice score (DSC) of the volume (not the slice). The third row depicts the subtraction map (error map) between the target and the warped source image. The value is within the range [-1,1] since the image intensities are normalized to [0,1]. The last row shows the warped grid of the deformation field. The two numbers shown are the mean foreground displacement magnitude and the percentage of foreground non-diffeomorphic voxels, both statistics are computed in the whole image volume.
    }
    \label{fig:oasis}
\vspace{-0.1cm}
\end{figure}

\begin{figure}[ht]
    \centering
    \includegraphics[width=\linewidth]{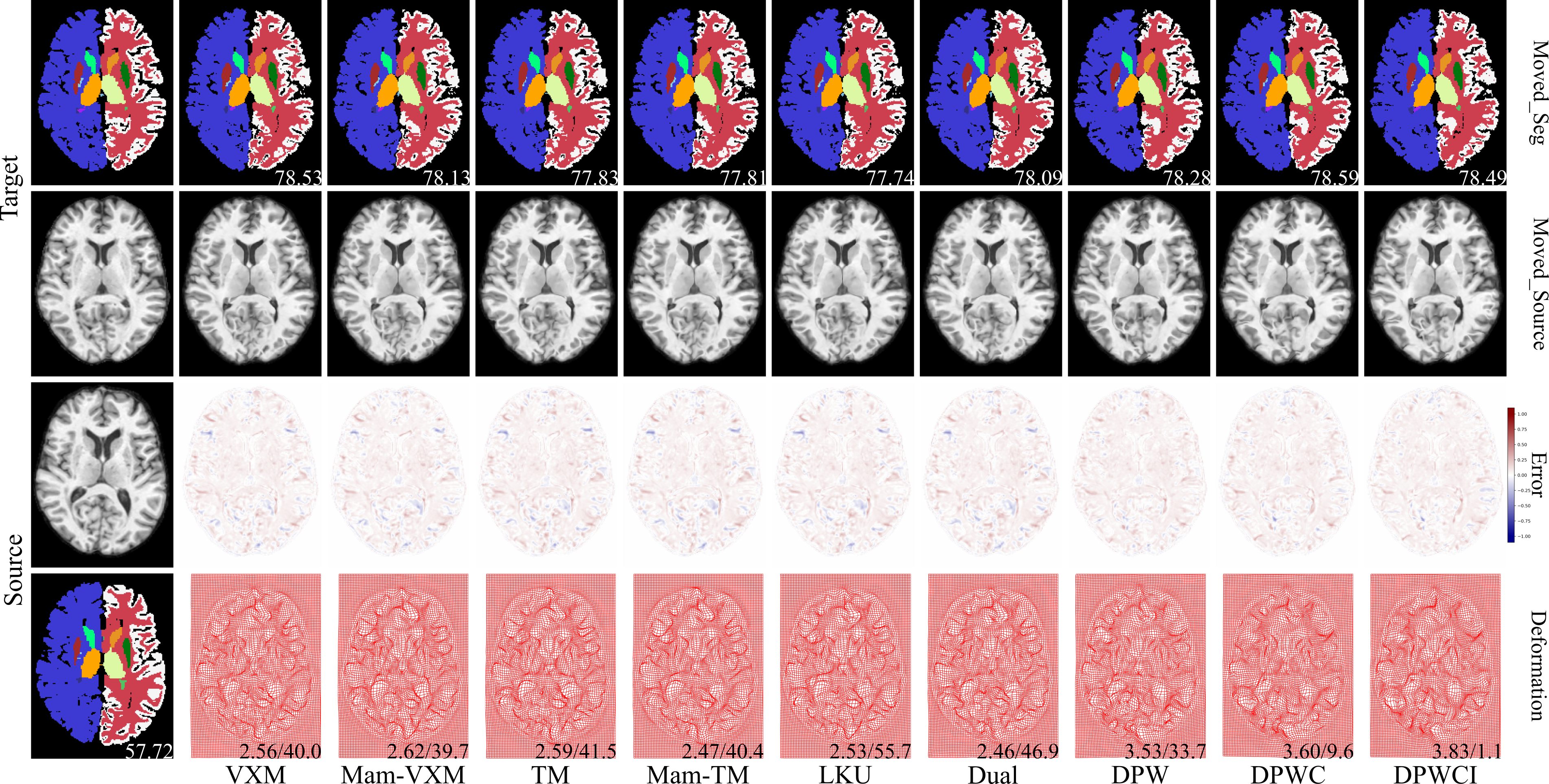}
    \suppfigure{
    The axial-viewed visualization result of a randomly sampled pair from the ADNI dataset. The first column shows the target label map and image, and the source image and label map from top to bottom. The other columns correspond to the methods to be compared. The first and second rows show the warped source (moved) label map and image by the methods. The bottom right number shown in the moved label map is the mean dice score (DSC) of the volume (not the slice). The third row depicts the subtraction map (error map) between the target and the warped source image. The value is within the range [-1,1] since the image intensities are normalized to [0,1]. The last row shows the warped grid of the deformation field. The two numbers shown are the mean foreground displacement magnitude and the percentage of foreground non-diffeomorphic voxels, both statistics are computed in the whole image volume.
    }
    \label{fig:adni}
\vspace{-0.1cm}
\end{figure}

\begin{figure}[ht]
    \centering
    \includegraphics[width=\linewidth]{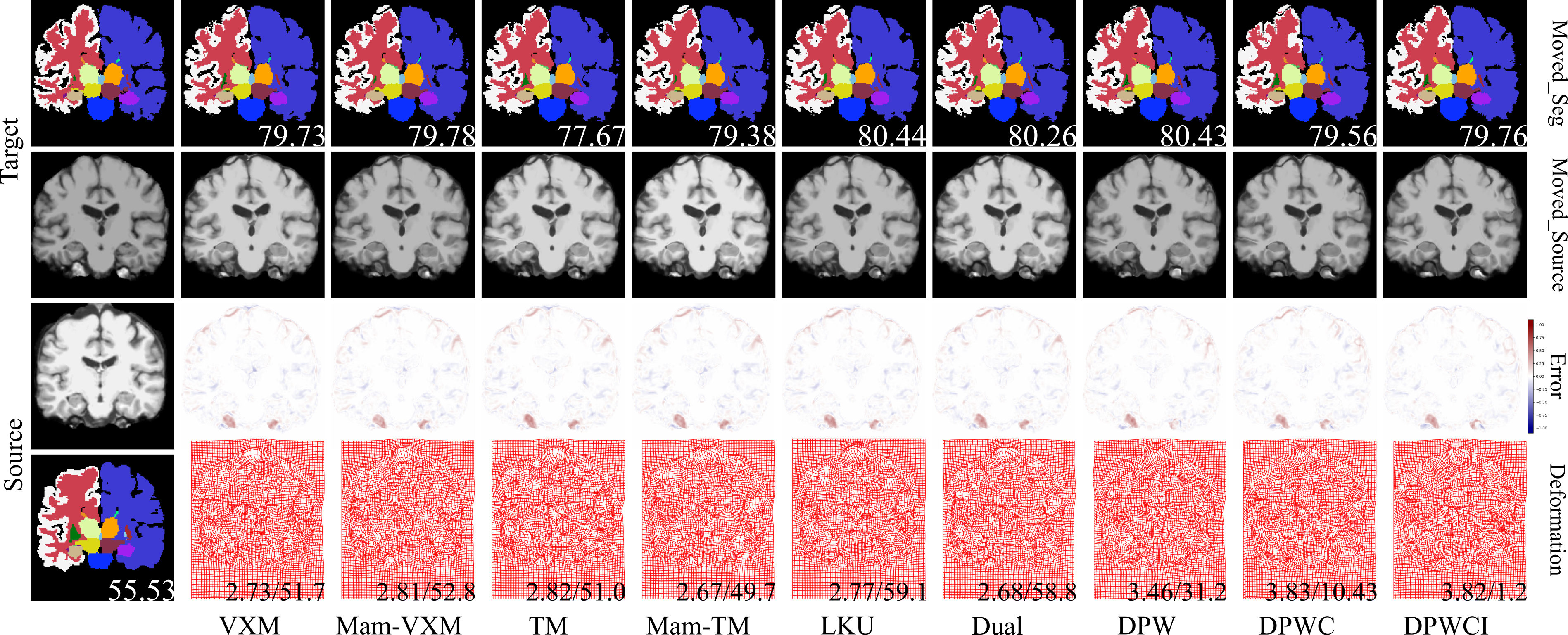}
    \suppfigure{
    The coronal-viewed visualization result of a randomly sampled pair from the IXI dataset. The first column shows the target label map and image, and the source image and label map from top to bottom. The other columns correspond to the methods to be compared. The first and second rows show the warped source (moved) label map and image by the methods. The bottom right number shown in the moved label map is the mean dice score (DSC) of the volume (not the slice). The third row depicts the subtraction map (error map) between the target and the warped source image. The value is within the range [-1,1] since the image intensities are normalized to [0,1]. The last row shows the warped grid of the deformation field. The two numbers shown are the mean foreground displacement magnitude and the percentage of foreground non-diffeomorphic voxels, both statistics are computed in the whole image volume.
    }
    \label{fig:ixi}
\vspace{-0.1cm}
\end{figure}


\begin{table}
\supptable{Cross-sectional registration results of brain T1-MRI. 200 pairs are randomly sampled from 84 subjects in OASIS. SD$\log$J is in the scale $\times 10^2$.}\label{tab:oasis}
    \centering
    \setlength{\tabcolsep}{2.5pt} 
\begin{tabular}{lcccc}
    \toprule
    & \multicolumn{4}{c}{OASIS}
    \\
    \cmidrule(lr){2-5}
       &  DSC $\uparrow$ & HD90 $\downarrow$ & SD$\log$J $\downarrow$ & NDV(\%) $\downarrow$\\
    \midrule
    affine & 57.3$\pm$7.0 & - & - & - \\
    Pool-Up & 84.0$\pm$2.3 & 4.94$\pm$0.83 & 9.01$\pm$1.08 &0.43$\pm$0.14 \\
    VXM & 84.3$\pm$2.2 & 4.91$\pm$0.82 & 9.27$\pm$1.31 &0.50$\pm$0.14 \\
    Mam-VXM & 84.8$\pm$2.0 & 4.85$\pm$0.84 & 9.41$\pm$1.43 &0.51$\pm$0.15 \\
    TM & 83.7$\pm$2.1 & 4.89$\pm$0.80 & 9.51$\pm$0.95 &0.56$\pm$0.13 \\
    Mam-TM & 83.5$\pm$2.1 & 4.93$\pm$0.82 & 9.44$\pm$1.16 &0.54$\pm$0.15 \\
    LKU & 84.7$\pm$2.0 & 4.84$\pm$0.82 & 9.75$\pm$1.43 &0.62$\pm$0.18 \\
    Dual & 84.5$\pm$2.1 & 4.87$\pm$0.78 & 9.71$\pm$1.25 &0.62$\pm$0.18 \\
    VXM-P & 83.5$\pm$2.3 & 4.91$\pm$0.84 & 9.14$\pm$1.02 &0.52$\pm$0.14 \\
    DP & 83.2$\pm$2.1 & 4.91$\pm$0.80 & 8.87$\pm$0.78 &0.55$\pm$0.15 \\
    DWP & 85.0$\pm$1.8 & 4.86$\pm$0.78 & 8.85$\pm$0.80 &0.41$\pm$0.07 \\
    DWPI & 84.7$\pm$1.8 & 4.90$\pm$0.78 & 8.74$\pm$0.53 &0.04$\pm$0.02 \\
    DWCP & 85.4$\pm$1.7 & 4.88$\pm$0.81 & 8.09$\pm$0.61 &0.13$\pm$0.04 \\
    DWCPI & 85.6$\pm$1.7 & 4.83$\pm$0.77 & 8.62$\pm$1.03 &0.02$\pm$0.02 \\
    \bottomrule
\end{tabular}
\end{table}

\begin{table}
\supptable{Cross-sectional registration results of brain T1-MRI. 200 pairs are randomly sampled from 43 subjects in ADNI. SD$\log$J is in the scale $\times 10^2$.}\label{tab:adni}
    \centering
    \setlength{\tabcolsep}{2.5pt} 
\begin{tabular}{lcccc}
    \toprule
    & \multicolumn{4}{c}{ADNI}
    \\
    \cmidrule(lr){2-5}
       &  DSC $\uparrow$ & HD90 $\downarrow$ & SD$\log$J $\downarrow$ & NDV(\%) $\downarrow$\\
    \midrule
    affine & 52.8$\pm$5.1 & - & - & - \\
    Pool-Up & 77.9$\pm$1.9 & 4.35$\pm$0.62 & 9.23$\pm$0.68 &0.50$\pm$0.12 \\
    VXM & 78.1$\pm$1.8 & 4.36$\pm$0.60 & 9.10$\pm$0.56 &0.50$\pm$0.12 \\
    Mam-VXM & 78.3$\pm$1.8 & 4.31$\pm$0.58 & 9.45$\pm$0.72 &0.56$\pm$0.14 \\
    TM & 77.6$\pm$1.8 & 4.34$\pm$0.60 & 9.54$\pm$0.66 &0.57$\pm$0.14 \\
    Mam-TM & 77.6$\pm$1.8 & 4.37$\pm$0.61 & 9.49$\pm$0.73 &0.57$\pm$0.16 \\
    LKU & 78.3$\pm$1.7 & 4.30$\pm$0.60 & 10.14$\pm$1.19 &0.70$\pm$0.20 \\
    Dual & 77.9$\pm$2.0 & 4.35$\pm$0.62 & 9.52$\pm$0.71 &0.63$\pm$0.13 \\
    VXM-P & 77.4$\pm$1.9 & 4.38$\pm$0.60 & 9.04$\pm$0.63 &0.52$\pm$0.13 \\
    DP & 76.7$\pm$2.0 & 4.40$\pm$0.61 & 8.79$\pm$0.42 &0.55$\pm$0.14 \\
    DWP & 78.8$\pm$1.7 & 4.35$\pm$0.60 & 8.75$\pm$0.42 &0.43$\pm$0.06 \\
    DWPI & 78.5$\pm$1.7 & 4.27$\pm$0.60 & 8.53$\pm$0.39 &0.04$\pm$0.02 \\
    DWCP & 79.2$\pm$1.7 & 4.27$\pm$0.61 & 7.89$\pm$0.35 &0.14$\pm$0.02 \\
    DWCPI & 79.4$\pm$1.7 & 4.22$\pm$0.59 & 8.14$\pm$0.36 &0.01$\pm$0.00 \\
    \bottomrule
\end{tabular}
\end{table}

\begin{table}
\supptable{Cross-sectional registration results of brain T1-MRI. 200 pairs are randomly sampled from 115 subjects in IXI. SD$\log$J is in the scale $\times 10^2$.}\label{tab:ixi}
    \centering
    \setlength{\tabcolsep}{2.5pt} 
\begin{tabular}{lcccc}
    \toprule
    & \multicolumn{4}{c}{IXI}
    \\
    \cmidrule(lr){2-5}
       &  DSC $\uparrow$ & HD90 $\downarrow$ & SD$\log$J $\downarrow$ & NDV(\%) $\downarrow$\\
    \midrule
    affine & 54.5$\pm$4.9 & - & - & - \\
    Pool-Up & 76.5$\pm$2.6 & 4.94$\pm$0.78 & 9.21$\pm$0.62 &0.48$\pm$0.15 \\
    VXM & 76.8$\pm$2.4 & 4.96$\pm$0.78 & 9.33$\pm$0.64 &0.52$\pm$0.15 \\
    Mam-VXM & 76.9$\pm$2.4 & 4.96$\pm$0.76 & 9.55$\pm$0.79 &0.53$\pm$0.15 \\
    TM & 76.2$\pm$2.3 & 4.97$\pm$0.77 & 9.60$\pm$0.67 &0.57$\pm$0.17 \\
    Mam-TM & 76.3$\pm$2.3 & 4.94$\pm$0.78 & 9.53$\pm$0.62 &0.57$\pm$0.18 \\
    LKU & 77.1$\pm$2.3 & 4.93$\pm$0.75 & 9.89$\pm$0.90 &0.63$\pm$0.18 \\
    Dual & 76.7$\pm$2.4 & 4.94$\pm$0.78 & 9.62$\pm$0.78 &0.61$\pm$0.17 \\
    VXM-P & 76.0$\pm$2.4 & 5.00$\pm$0.74 & 9.14$\pm$0.66 &0.52$\pm$0.15 \\
    DP & 75.9$\pm$2.3 & 4.97$\pm$0.77 & 9.03$\pm$0.52 &0.56$\pm$0.18 \\
    DWP & 77.2$\pm$2.3 & 4.90$\pm$0.73 & 8.91$\pm$0.59 &0.35$\pm$0.07 \\
    DWPI & 76.9$\pm$2.2 & 4.87$\pm$0.74 & 9.11$\pm$0.85 &0.07$\pm$0.07 \\
    DWCP & 77.5$\pm$2.3 & 4.86$\pm$0.75 & 8.23$\pm$0.51 &0.14$\pm$0.05 \\
    DWCPI & 77.8$\pm$2.2 & 4.85$\pm$0.74 & 8.47$\pm$0.48 &0.02$\pm$0.01 \\
    \bottomrule
\end{tabular}
\end{table}

\end{document}